\documentclass[11pt, a4paper, logo, copyright]{deepmind}

\pdftrailerid{redacted}

\makeatletter
\renewcommand\bibentry[1]{\nocite{#1}{\frenchspacing\@nameuse{BR@r@#1\@extra@b@citeb}}}
\makeatother

\usepackage{kantlipsum, lipsum}
\usepackage{dm-colors}
\usepackage{dsfont}
\usepackage{float}
\usepackage{listings}

\usepackage[authoryear, sort&compress, round]{natbib} 

\usepackage{algpseudocode}

\definecolor{mygreen}{rgb}{0,0.6,0}
\definecolor{mymauve}{rgb}{0.58,0,0.82}

\graphicspath{{figures/}}

\title{Using Unity to Help Solve Intelligence}

\correspondingauthor{tomward@google.com}

\reportnumber{} %

\renewcommand{\today}{November 2020}

\author[*,1]{Tom Ward}
\author[*,1]{Andrew Bolt}
\author[*,1]{Nik Hemmings}
\author[1]{\\ Simon Carter}
\author[1]{Manuel Sanchez}
\author[1]{Ricardo Barreira}
\author[1]{Seb Noury}
\author[1]{Keith Anderson}
\author[1]{Jay Lemmon}
\author[1]{Jonathan Coe}
\author[1]{Piotr Trochim}
\author[1]{Tom Handley}
\author[1]{Adrian Bolton}

\affil[*]{Equal contributions}
\affil[1]{DeepMind Technologies, London}

\banner{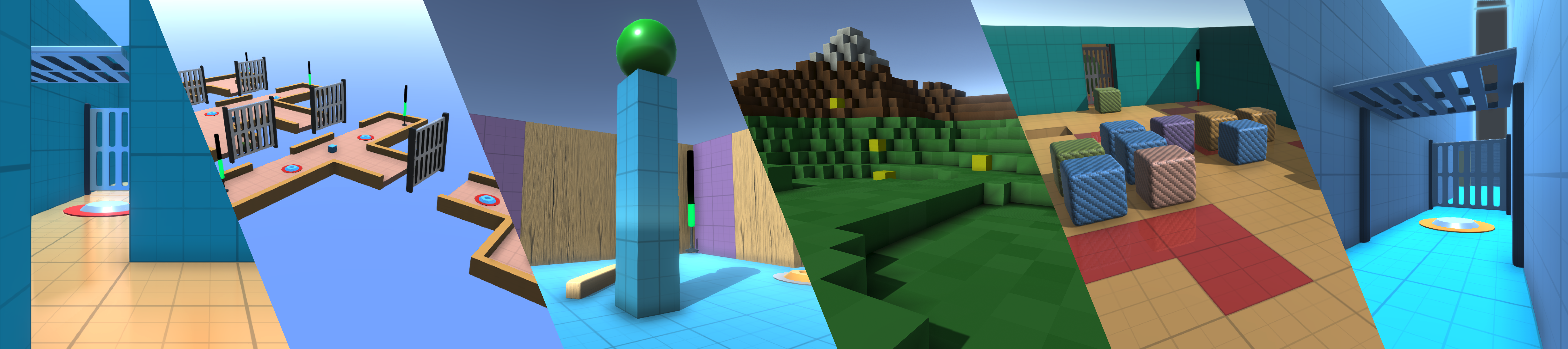}{A selection of Unity-based environments built using our approach.}

\begin{abstract}
In the pursuit of artificial general intelligence, our most significant measurement of progress is an agent's ability to achieve goals in a wide range of environments.
Existing platforms for constructing such environments are typically constrained by the technologies they are founded on, and are therefore only able to provide a subset of scenarios necessary to evaluate progress.
To overcome these shortcomings, we present our use of Unity, a widely recognized and comprehensive game engine, to create more diverse, complex, virtual simulations.
We describe the concepts and components developed to simplify the authoring of these environments, intended for use predominantly in the field of reinforcement learning.
We also introduce a practical approach to packaging and re-distributing environments in a way that attempts to improve the robustness and reproducibility of experiment results.
To illustrate the versatility of our use of Unity compared to other solutions, we highlight environments already created using our approach from published papers.
We hope that others can draw inspiration from how we adapted Unity to our needs, and anticipate increasingly varied and complex environments to emerge from our approach as familiarity grows.
\end{abstract}

\begin{document}
\maketitle

\section{Introduction}
\label{section:introduction}

In the past decade, reinforcement learning has made significant advances in the pursuit of achieving general purpose intelligence \citep[e.g.][]{mnih2015, silver2016, espeholt2018impala}.
A widely accepted definition of the term general intelligence is ``an agent's ability to achieve goals in a wide range of environments'' \citep{legg2007universal}.
As such, there has been an increase in the creation of virtual environments, both to probe specific cognitive abilities and to benchmark an agent's performance.
Games have long been used as a means to evaluate progress in artificial intelligence, beginning with board games such as backgammon \citep{tesauro1994td}, chess \citep{campbell2002deep} or Go \citep{silver2016}.
More recently, video games have also proved a rich source of learning environments.
Popular examples include the Arcade Learning Environment \citep{bellemare2013arcade}, the Retro Learning Environment \citep{bhonker2016playing, nichol2018gotta}, multiplayer first-person game Quake III \citep{jaderberg2019human}, as well as real-time strategy games Dota 2 \citep{berner2019dota} and StarCraft II \citep{vinyals2017starcraft}.
Both board and video games prove popular because they provide a rich source of synthetic observational data, typically have a clear measure of success, progress can be reviewed empirically, and agent performance can be directly evaluated against human performance. Additionally, such games have an extra measure of validity because they are not defined by AI researchers, and are not designed specifically for the purpose of AI research.

As demand for ever more complex environments increased, so did the engineering effort required to transform existing interactive games into useful research environments.
Consequently, there has been a paradigm shift towards using a video game's underlying engine as a platform for creating larger suites of environments.
Examples include VizDoom \citep{Kempka2016ViZDoom}, DeepMind Lab \citep{beattie2016deepmind} and Malmo \citep{johnson2016malmo}, which use the underlying engines of popular video games Doom, Quake III and Minecraft respectively.
Whilst these platforms have been instrumental in advancing progress, the games they are founded upon each have design constraints that restrict their potential for environment diversity.
For example, DeepMind Lab proved an excellent platform for exploring complex navigational tasks \citep{banino2018} and classical laboratory psychological experiments \citep{leibo2018psychlab}, among others, but is limited to first-person perspectives and primitive physics. This restricts the scope of the platform to a relatively narrow subset of possible experiments, and affects its applicability for evaluating general intelligence.
At the same time, the video games industry has also evolved so that the majority of games are now created using more general-purpose engines, which were developed from the outset to enable a wide variety of games to be built upon their core technologies.

To enable us to create a wide range of reinforcement learning environments \citep{sutton2018reinforcement}, we chose to use Unity \citep{unity3d}, a popular commercial game engine.
Unity's flexibility in visual and simulation fidelity, programmer-friendly tools, and large, active community makes it an excellent choice as a foundation for building our learning environments.

Another important consideration made whilst devising our approach is the question of reproducibility, defined as ``the ability of a researcher to duplicate the results of a prior study using the same materials and procedures as were used by the original investigator'' \citep{bollen2015social}.
This has become a pressing subject in several scientific fields \citep{baker20161, dodge2019show}, including within the reinforcement learning community \citep{henderson2017deep, pineau2020improving}.

In this paper, we describe the high-level concepts developed to adapt Unity to our needs.
We detail the means by which we interact with these environments through an open-sourced communication protocol.
We present a practical approach to packaging a pre-built environment and all its dependencies in an easily distributable form.
Finally, we evaluate the success of our solution by highlighting several papers published using this approach.
\section{Related Work}

Creating reinforcement learning environments using Unity has some precedent.
Indeed, the creators of Unity have themselves released a plugin: Unity ML-Agents toolkit \citep{juliani2020unity}.
This is a ready-to-use toolkit that includes a variety of learning algorithms, training regimes and sample environments, as well as a mechanism to connect externally-defined agents.
The toolkit focuses on simplifying the process of making an existing Unity project available as a learning environment, and subsequently producing trained agents.
A variety of game development applications are anticipated for agents trained with this toolkit, such as controlling non-player characters, enabling richer automated testing, or aiding new game design decisions.
Their toolkit also supplies a Python API to allow externally defined agents to interact with configured games.

Whilst there are many similarities between our approach and the Unity ML-Agents toolkit, the solutions differ in their primary intents.
The focus of our work is to make it as simple as possible to construct a heterogeneous set of environments, and package these up as configurable black-box servers that any number of agents can connect to.
This strong separation of concerns between environments and agents aims to encourage research autonomy, allowing advances in learning approaches to take place independently of simulation authoring.
We trade this against having a simple method of embedding pre-trained agents directly into the environment, for which our approach may not be best suited.
Our decoupling approach has several other benefits, such as distributed training, persistent simulations, simplified multi-agent support, and better guarantees of experiment reproducibility.
It is important to note that both propositions have equal merit with respect to their overall objective.

As well as the Unity ML-Agents toolkit, several other platforms have been developed to enable the creation of complex environments.
As mentioned in Section~\ref{section:introduction}, most of these platforms \citep{beattie2016deepmind, Kempka2016ViZDoom, johnson2016malmo} leverage existing video game engines.
Others, such as DeepMind Control Suite \citep{deepmindcontrolsuite2018}, utilize specialist engines such as MuJoCo, a high quality physics engine.
While each of these platforms provide some degree of environment customization, they are also dependent on highly-specialized engines, originally built for a specific set of needs.
This specialization means environments created within these platforms are inherently constrained by the underlying engine.
For instance, MuJoCo-based environments are able to accurately simulate multi-joint dynamics, but are less capable at producing visually accurate scene representations.
In some cases, this specialization is desirable as it can ensure a certain level of consistency.
For example, using Malmo \citep{johnson2016malmo}---a platform based on the video game Minecraft---there is an intrinsic uniformity to how the world is represented, and thus how an agent can interact with the world.
This can ground an agent's learning, simplifying the transfer of knowledge from one task to another.
\cite{juliani2020unity} provides a more in-depth analysis of these and other platforms.
We consider the versatility of using Unity, and its suitability for creating a wider variety of tasks, as an acceptable trade-off against this implicit consistency.

There have also been attempts to improve the reproducibility of scientific research, by re-purposing solutions in the field of software engineering \citep{tanner2009rl}.
One such solution is to use ``Docker'' \citep{boettiger2015introduction}, which provides a mechanism for packaging up a piece of software to be reliably run from one computing environment to another.
There is precedent for using Docker within the realm of AI research.
An example of this is OpenAI Universe \citep{openaiuniverse}, a platform where environments are packaged using Docker and agents interact through a remote desktop protocol (VNC) and auxiliary connection.
We propose using a similar approach, but instead of using a generic protocol such as VNC, we use a more bespoke, well-defined means of communication.
This has the benefit of being more efficient, practical and reliable than using generic protocols.

\section{Technical Details}

To create the large set of heterogeneous environments necessary to advance artificial intelligence, we ground our solution in Unity.
Unity can be considered as two separate parts: a cross-platform runtime environment known as the `Player', and a content creation tool known as the `Editor'.
Using the Unity Editor, we produce new environments by creating a Unity project with one or more \texttt{Scene} objects.
Each \texttt{Scene} represents a different initial world state for the simulation.

A \texttt{Scene} consists of a hierarchy of \texttt{GameObject} entities.
Each \texttt{GameObject} represents a physical or logical item in the environment, and its behaviour is determined by the set of components that are assigned to it.
Unity provides many common components as standard.
For example \texttt{Mesh} components for assigning geometry, \texttt{RigidBody} components to control an object's position via a physics simulator, or \texttt{Renderer} components to determine how to draw the \texttt{GameObject} to the screen.
Importantly, it is also possible to create new components by utilizing Unity's C\# scripting API.

\subsection{Overview}

\begin{figure}
    \centering
    \includegraphics[scale=0.45]{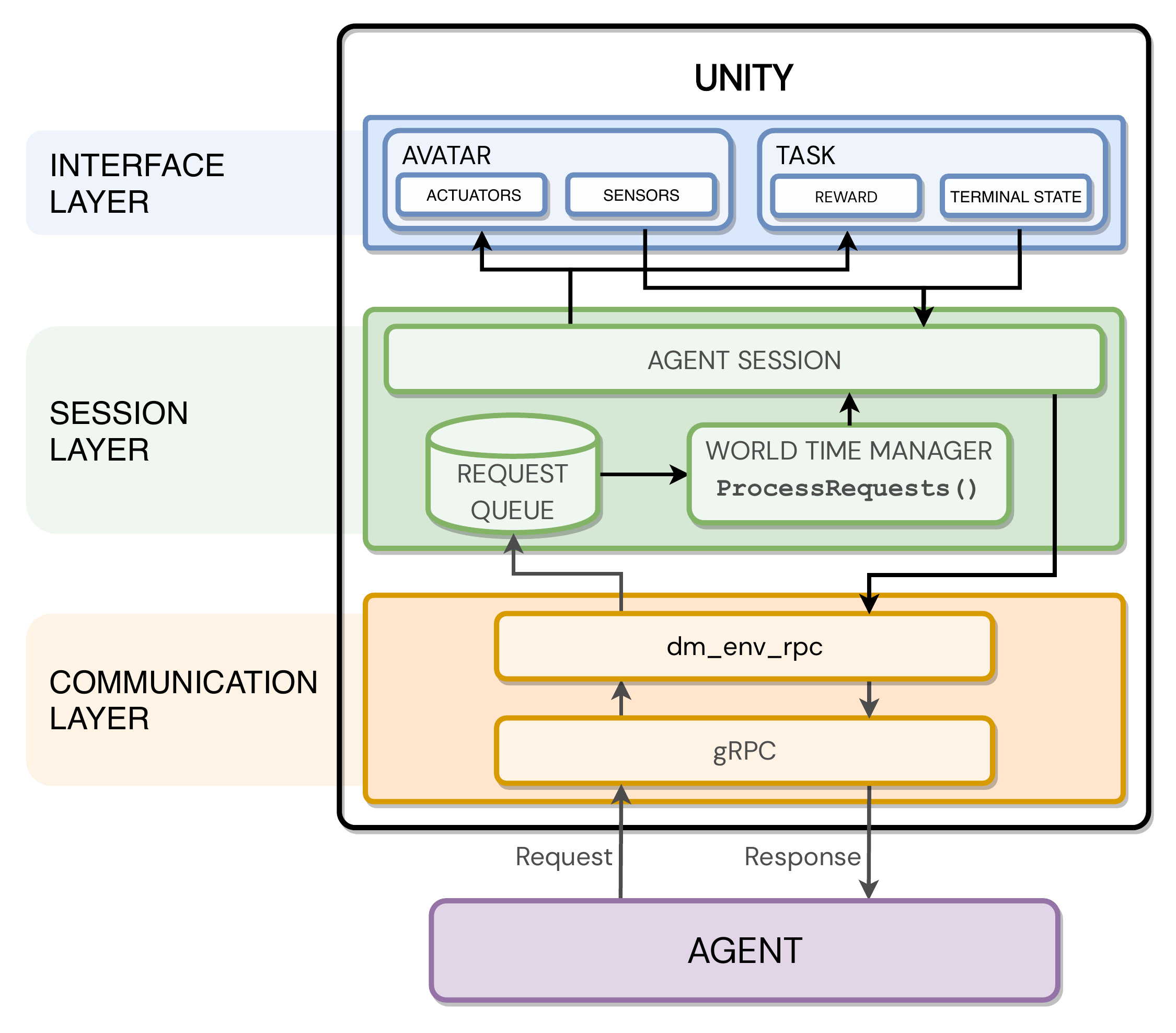}
    \caption{High-level overview of our layers and their interactions. For more details on the Interface, Session and Communication layers, see sections \ref{section:interface_layer}, \ref{section:session_layer} and \ref{section:communication_layer} respectively.}
    \label{fig:dmunity_diagram}
\end{figure}

We now consider the problem of exposing these environments to learning agents.
From a high-level perspective, our approach can be described as three inter-connected layers: the interface, session, and communication layers.
Each layer is responsible for a specific stage of the agent-environment interaction, starting with transforming a Unity project into a reinforcement-learning environment, through to the low-level communication protocol necessary to communicate with one or more agents.
Figure~\ref{fig:dmunity_diagram} provides a broad overview of each layer and how they interrelate.

\subsection{Interface Layer}
\label{section:interface_layer}

The purpose of the interface layer is to simplify the process of converting a Unity project into a learning environment.
To achieve this, an environment creator needs to specify the following:
\begin{samepage}
\begin{itemize}
    \item The set of actions a connected agent is permitted to use to interact with an environment.
    \item The set of observations an agent can request from the environment.
    \item The environment's operating conditions, for example how and when to advance time within the simulation.
\end{itemize}
\end{samepage}
Concretely, our approach addresses this through the introduction of two concepts: \emph{Avatars} and \emph{Tasks}.

\subsubsection{Avatars, Actuators and Sensors}

Agents interact with the simulation by taking control of an \emph{Avatar}, which is typically represented by a \texttt{GameObject}.
The job of an Avatar is to provide a collection of \emph{Actuators} and \emph{Sensors}.
Actuators provide the means by which an agent can influence the simulation, while Sensors are how are an agent can observe the simulation.
An Avatar is typically a single physical embodiment, but it doesn't necessarily need to be---it could be something abstract like the cooling system for a factory, or a spectator that is able to observe certain characteristics of a simulation.

We use C\# attributes to tag fields on component objects as Actuators or Sensors.
These components are then attached to the Avatar's \texttt{GameObject}, which are subsequently aggregated by scanning the \texttt{GameObject} for these attributes to create a final set of Actuators and Sensors.
We also allow the inclusion of auxiliary information through the use of attribute parameters.
Examples of which include a means to customize the name of an Actuator or Sensor, to define an Actuator's valid range, or relevant Sensor metadata.
Listing~\ref{code:actuator} shows an example that exposes two Actuators: \texttt{MOVE\_BACK\_FORWARD} which has a valid range of $(-1:1)$, and a Boolean Actuator named \texttt{JUMP}.

\begin{minipage}{0.95\linewidth}
\begin{lstlisting}[language={[Sharp]C}, caption={Example Avatar scalar Actuators.}, captionpos=b, label={code:actuator}]
[ScalarActuator("MOVE_BACK_FORWARD", Min = -1.0f, Max = 1.0f)]
public float MoveBackForward;

[ScalarActuator("JUMP")]
public bool Jump;
\end{lstlisting}
\end{minipage}

Support is provided for all primitive types, with the Sensor's dimensions inferred either from the type, or explicitly defined through a \texttt{Shape} attribute parameter.
We also provide support for more complex Unity types, the most prominent of which are Unity Cameras.
When an agent requests observations from a Camera Sensor type, we ask Unity to render the Camera's view out to a multi-dimensional array of bytes.
Listing~\ref{code:sensor} shows four examples of different Sensor types that can be observed by an agent:

\begin{minipage}{0.95\linewidth}
\begin{lstlisting}[language={[Sharp]C}, 
    caption={Four example Avatar Sensors: A scalar \texttt{SCORE}, 1-dimensional \texttt{ACCELERATION}, 2-dimensional \texttt{TRANSFORM} and 3-dimensional Camera \texttt{PIXELS}.}, 
    captionpos=b,
    label={code:sensor}]
[Sensor("SCORE")]
public float Score;

[Sensor("ACCELERATION")]
public Vector3 Acceleration;

[Sensor("TRANSFORM", Shape = new int[] { 3, 4 })]
public double[] TransformMatrix = new double[] { 1.0, 0.0, 0.0, 4.0,
                                                 0.0, 1.0, 0.0, 5.0,
                                                 0.0, 0.0, 1.0, 6.0 };

[CameraSensor("PIXELS", CameraObservation.Format.RGB24,
              DefaultWidth = "96", DefaultHeight = "72")]
public Camera RgbCameraSensor;
\end{lstlisting}
\end{minipage}

\subsubsection{Tasks}

The purpose of a \emph{Task} is to define an agent's operating conditions whilst in the environment.
These include:
\begin{itemize}
    \item \textbf{What Avatar the agent controls.} \\
    A Task can either spawn a new Avatar \texttt{GameObject}, or take possession of one that already exists.
    \item \textbf{The agent's current reward.} \\
    Rewards are exposed by the Task as a single scalar value, discretely evaluated once per step which is typical for reinforcement learning \citep{sutton2018reinforcement}.
    It's also possible to derive more complex reward signals using the output of Avatar Sensors if necessary.
    \item \textbf{How to start a new episode.} \\
    A Task implements a \texttt{StartEpisode\(\)} function to begin a new episode.
    At the start of an episode, the simulation is also typically reset to a state drawn from a stationary distribution.
    \item \textbf{The conditions under which an episode can terminate.} \\
    When a terminal state is reached, the Task can be queried for the type of episode termination.
    This enables agents to disambiguate whether the episode reached a terminal state or if, for example, a predetermined time limit was reached.
    \item \textbf{How and when to advance time within the simulation.} \\
    Tasks may also need to manipulate the agent's perception of time.
    For example, we may wish to run a physics simulation for a few seconds as preparation for starting a new episode.
    See Section~\ref{section:time_stepping} for more details on stepping time.
\end{itemize}
By separating the Task logic from the \texttt{Scene} content used to construct the simulated world, we can mix and match Tasks and content to create a variety of different scenarios for agents.
A Task can be further configured through settings provided by the agent.
We can also assign each connected agent a different Task.
As a result, it's conceivable for agents connected to the same simulation to experience different episode boundaries, and even different notions of passing time, although the most common multiplayer setting is for all agents to have their steps and episodes in lockstep.

\subsection{Session Layer}
\label{section:session_layer}

A \emph{Session} represents the current state of a single agent's connection to the environment.
It is responsible for responding to agent requests received from the communication layer.
When an agent connects, a \texttt{SessionFactory} creates a new Session based on the settings provided.
In practice, a Session is primarily responsible for amalgamating the Task and Avatar components.
For example, when an agent steps the environment, the Session forwards actions to the Avatar's Actuators, and aggregates the requested observations from the corresponding Avatar Sensors.
The Session is also responsible for handling other requests outside the standard reinforcement learning paradigm, such as environment diagnostic information, or to enable researchers to make controlled interventions to modify the simulation.

\subsubsection{Advancing Time}
\label{section:time_stepping}

Unity provides support for approximating continuous time as a series of discrete simulation frames.
Each frame can advance the simulation state by a fractional ``delta-time''.
In a typical game setting, Unity dynamically adjusts this delta-time for the next frame to maintain an illusion of the simulation running at the same speed as the real-world.
For instance, Unity will increase the amount of time between frames if the scene becomes too complex to simulate, reducing the simulation's fidelity.

For reproducible machine-learning results, this adaptive real-time frame rate behaviour can be undesirable.
To mitigate this behaviour, we switch Unity into a mode where we have explicit control over the time between frames.
Concretely, we achieve this by controlling Unity's \texttt{Time.captureFrameRate} and \texttt{Time.timeScale} attributes.
Note that agents are not required to observe every simulation frame.
It's the responsibility of the agent's Session to determine when to process agent actions, and how long to wait before returning the corresponding observations.
For example, in the Construction environment (see Section~\ref{section:results} for more details), each agent action places a new block in the environment.
Before the agent receives the next observation, the environment advances $N$ frames to simulate a ball rolling over the placed blocks, returning the observations once the ball has reached its subsequent resting state.
The result is that the agent observes a single action/observation pair, yet the environment has advanced many frames.

We co-ordinate the environment and connected agents through a single, global object called the \emph{World Time Manager} (WTM).
The WTM serves two main functions.
Firstly, it provides a centralized scheduling service for precise execution of each agent’s requests to their respective Sessions.
This ensures that agent requests are reliably processed at a specific point in the update loop, in between simulation state changes.
Secondly, the WTM manipulates the advancement of simulation time based on the outcomes from executing the scheduled agent requests.
To achieve this, our implementation requires that all agent requests are defined within the Session layer strictly in terms of operations on their Session, and that each request is required upon completion to return a simulation advancement preference.
We describe this preference as a \texttt{TickState}.

A \texttt{TickState} can be one of the following:
\begin{itemize}
    \item \texttt{MustTick}: The simulation must advance before the next action is processed.
    \item \texttt{MustNotTick}: The simulation must not advance before the next action is processed.
    \item \texttt{MayTick}: The simulation may optionally choose to advance based on external criteria, such as reaching a maximum amount of time for processing agent actions.
\end{itemize}

Agent interactions are handled by worker threads. The agent controllers use thread-safe queues to schedule work to be performed by the WTM on a specific tick.
At the end of each tick, the WTM makes a ``pending'' list, representing agents that still have outstanding requests for the current simulation frame.
Next, the WTM processes each queue by sequentially sending agent requests to its Session and receiving the corresponding \texttt{TickState}.
If an action returns \texttt{MustTick}, the WTM transfers the remaining requests in the agent's queue to a new, ``future'' list.
Processing continues on all remaining queues until each queue either returns \texttt{MustTick}, or the agent's queue is empty.
At this point, the WTM allows Unity to run its update loop, after which the WTM transfers the future list to become the new pending list.

This approach allows all connected agents to be processing observations concurrently, while maintaining temporal consistency with their interactions with the simulation.

\subsection{Communication Layer}
\label{section:communication_layer}

The communication layer has two primary responsibilities: establishing a channel of communication between each agent and the environment, and marshalling the agent requests and responses to and from the session layer.

For the channel of communication, we use Google's high performance, open-source remote procedure call (RPC) framework, \emph{gRPC} \citep{grpc}.
Specifically, we use its bi-directional streaming RPC variant to send and receive ordered streams of messages from each connected agent.
gRPC's versatility to work with multiple runtime platforms and programming languages makes it a good fit for our agent to environment interoperability.

Next, we require a messaging convention to communicate.
To achieve this, we introduce \allowbreak\emph{dm\_env\_rpc} \citep{dm_env_rpc2019}, a messaging protocol intended to standardize the communication between machine learning agents and environments.
Each agent client has a single, bi-directional RPC stream to a Unity environment server.
A server can accept more than one simultaneous agent connection, depending on whether the simulation supports such a use-case.
Each server sends exactly one response for each agent request, where the message of the response corresponds to that of the request (e.g. a \texttt{StepRequest} would yield a \texttt{StepResponse}), or an error message.
In a typical single-agent scenario, an agent would interact as follows:
\begin{enumerate}
    \item The agent sends a \texttt{CreateWorldRequest}, which sets up the Unity simulation conditioned on the request's settings (e.g. which Unity Scene to load).
    \item The agent next sends a \texttt{JoinWorldRequest} with an optional set of agent-centric settings.
    These might include which kind of Avatar to instantiate, or what team to join in a multi-agent setting.
    The \texttt{JoinWorldResponse} returns what actions and observations an agent can use to interact with the environment.
    \item The agent is now able to send $N$ \texttt{StepRequest} messages to step the environment.
    Each \texttt{StepRequest} includes a sparse set of actions to apply, and the observations to expect in response.
    A \texttt{StepResponse} is populated with these requested observations, retrieved from the agent's Avatar Sensors, and whether the agent's Task has reached a terminal state.
    \item The agent can at any time send a \texttt{ResetRequest} to start a new Task episode, or a \texttt{ResetWorld\-Request} to set the simulation back to its initial state.
\end{enumerate}
For more information on these and other requests that an agent can send, see the dm\_env\_rpc documentation at \url{http://github.com/deepmind/dm_env_rpc}.

\subsection{Agent API}

Whilst it is possible for agents to interact with environments directly through dm\_env\_rpc, the protocol is intended primarily to describe the on-the-wire protocol format, and as such favors performance over usability.
To simplify the interchange for agents, dm\_env\_rpc includes implementations of popular reinforcement learning interfaces.
Once such API is \emph{dm\_env} \citep{dm_env2019}, a Python interface already used in environments such as DeepMind's Control Suite \citep{deepmindcontrolsuite2018}.
dm\_env exposes a subset of the dm\_env\_rpc protocol, providing functionality in a way familiar to Python developers.
For further details on the dm\_env API, refer to the documentation at \url{http://github.com/deepmind/dm_env}.

\subsection{Reproducibility}

To best ensure the robust reproduction of results with our environments, we propose the use of ``containers'' to work in tandem with our aforementioned communication layer.
A container is ``a standard unit of software that packages up code and all its dependencies so the application runs quickly and reliably from one computing environment to another'' \citep{what_is_docker}.
Unlike traditional Virtual Machines (VMs) that virtualize at the physical device level, containers are instead an abstraction at the application layer.
This means that containers are leaner, more performant and can share resources more readily than their VM counterparts.
Containers are all isolated from one another, allowing each environment to have their own distinct set of software, libraries and configuration files.
Containers are also composable, meaning it's easier to integrate into other systems when reproducing results.

To provide this container functionality, we use \emph{Docker}, a popular platform for containerization.
Docker is supported on a wide range of operating systems and cloud platforms, ensuring our Unity environment is able to run in a multitude of locations.
It supports versioning, which means that historical versions of the same environment can run concurrently.
Docker also supplies technologies to further simplify the packaging and running of software applications.
For example, it provides a means of creating new Docker images by creating a text file, called a ``Dockerfile''.
This is a simple script--similar to a Makefile--that summarizes the steps needed to assemble the image and run the application.
Not only does this trivialize the creation of new images, it also provides a human readable summary of the dependencies, environment variables and arguments needed to start the environment.

Listing~\ref{lst:dockerfile} shows a Dockerfile example for a Unity application called ``dm\_unity\_examples''.
The example uses Docker's reserved, minimal image \texttt{scratch}, which indicates that the image does not depend on any base image, copying in the contents of a \texttt{dm\_unity\_examples} directory.
It then sets a couple of environment variables to define the renderer to use (see Section~\ref{section:rendering} for more information on rendering) before running the Unity application.

\lstdefinelanguage{Dockerfile}
{
  morekeywords={FROM, RUN, ENV, ADD, COPY, ENTRYPOINT},
  morecomment=[l]{\#},
  morestring=[b]"
}
\begin{minipage}{0.95\linewidth}
\begin{lstlisting}[language=Dockerfile, caption={Dockerfile example for a Unity application named dm\_unity\_examples.},
    label={lst:dockerfile}, captionpos=b]
FROM scratch
ADD dm_unity_examples /app/dm_unity_examples

ENV UNITY_RENDERER=software
ENV UNITY_OSMESA_PATH=/app/dm_unity_examples/external_libosmesa_llvmpipe.so
ENTRYPOINT [ "/app/dm_unity_examples/Linux64Player", \
    # Unity command-line flags.
    "-logfile",                                      \
    "-batchmode",                                    \
    "-noaudio",                                      \
    # Other command-line flags.
    "--uri_address=[::]:10000"                       \
]
\end{lstlisting}
\end{minipage}

Using Docker to package and run the environment, gRPC to establish a connection, dm\_env\_rpc to communicate and dm\_env to simplify the agent's environment interactions, we have all the components required to construct a robust and reliable means of connecting agents and environments.
Listing~\ref{lst:python_example} provides a reference implementation of how an agent can start an example Docker container, connect and send actions until the episode ends.

\begin{minipage}{0.98\linewidth}
\begin{lstlisting}[language=python,caption={Python example for creating, connecting and interacting with a containerized environment.}, label={lst:python_example}, captionpos=b]
import dm_env_rpc
import docker
import numpy as np

from dm_env_rpc.v1 import connection as dm_env_rpc_connection
from dm_env_rpc.v1 import dm_env_adaptor

container = docker.from_env().containers.run(
    "dm_unity_examples", detach=True, ports={10000:10000})

connection = dm_env_rpc_connection.create_secure_channel_and_connect(
    "localhost:10000")

env, _ = dm_env_adaptor.create_and_join_world(
    connection, create_world_settings={}, join_world_settings={})

time_step = env.reset()
while not time_step.last():
  action = {}
  for name, spec in env.action_spec().items():
    action[name] = np.random.uniform(spec.minimum, spec.maximum, spec.shape)
  time_step = env.step(action)
\end{lstlisting}
\end{minipage}
\section{Performance}

Whilst the simplicity and versatility of our approach are of utmost importance, it's vital that it is also computationally efficient.
In traditional video games, developers strive to attain a consistently low per frame latency to achieve a smooth, real-time experience.
For machine learning environments, the goal is to instead produce the largest amount of simulation experience, with respect to the computational resources available.
To maximize this simulation throughput, we run as many instances of the environment in parallel as possible.
Agents exploit this parallelization by using distributed architectures \citep{nair2015massively}, scaling learning to as many machines as the agent can leverage \citep{espeholt2018impala}.

This paradigm shift from per-frame latency to total throughput, and from human to artificial players, has implications in the way we use Unity.
Our most prominent changes are as follows:
\begin{itemize}
    \item To allow the simulation to run faster than real-time, by removing synchronization points that are no longer necessary (e.g. coordinating when to refresh a display device).
    \item To encourage agents to run each Unity instance for as long as possible.
    This is so that we can amortize any start-up costs, such as loading 3D meshes or images.
    \item Counterintuitively, we disable multi-threading to reduce CPU contention when running multiple instances of Unity in parallel, and to avoid the computational overhead of thread synchronization.
\end{itemize}
Other changes include disabling Unity features that are unnecessary, and defaulting to running the simulation with a time delta of 33ms, equivalent to an update frequency of 30Hz.

\subsection{Rendering}
\label{section:rendering}

Typically game engines present pixel data by directly rendering to a physical display device.
More often than not, machine learning agents are run on computers that don't have such a device.
We solve this problem by adding support for rendering without the need for a physical display, otherwise known as ``headless rendering''.

As well as not having a display device, these machines may not have access to a GPU for hardware accelerated graphics, or may not wish to use this device for rendering.
To this end, we also provide support for rendering on CPU, using the open-source LLVMpipe software rasterizer, provided through Mesa3d  \citep{mesa3d}.
To improve performance, we again disable multi-threading to reduce CPU contention and threading overhead, but we also disable texture compression.
This is because CPU rendering lacks dedicated hardware for texture look-ups, so by disabling it we improve CPU rendering throughput at the cost of texture memory.

\subsection{Performance Results}

As environment performance is heavily dependent on the complexity of the simulation and the observations an agent requests, we provide results based on a representative reinforcement-learning task, called ``Seek Avoid''.
This task is a first-person, 3-dimensional environment where the agent's goal is to pick up as many apples as possible, whilst avoiding any lemons.
An agent is provided rendered pixel observations per-simulation step, with the episode ending after a fixed amount of time.

Table \ref{tab:performance} shows the total frame rate (frames/second) when running multiple concurrent Unity instances on our reference environment.
We capture the total frame rate by sending random actions to each environment and receiving the RGB pixel observations, rendered at a resolution of $96\times72$ which is typical for learning agents \citep{paine2019making}.
The results below were recorded on a Linux desktop with two 18-core Intel Xeon Gold 6154 3GHz CPUs and an NVIDIA Quadro P1000 GPU.

\begin{table}[H]
  \centering
  \begin{tabular}{l|ll|ll}
Unity Instances & CPU               &           & GPU               &           \\
                & Total frames/sec  & $\sigma$  & Total frames/sec  & $\sigma$  \\ \hline
1               & 445               & 4         & 1,060             & 54        \\
2               & 883               & 6         & 2,100             & 12        \\
4               & 1,791             & 2         & 4,129             & 63        \\
8               & 3,601             & 16        & 6,933             & 36        \\
16              & 6,788             & 34        & 9,982             & 15
  \end{tabular}
  \caption{Comparison of the total frame rates (frames/second) when running the Seek Avoid environment, rendering at a resolution of $(96\times72)$ on either CPU or GPU, with multiple concurrent instances of Unity.}
  \label{tab:performance}
\end{table}
\section{Results}
\label{section:results}

To evaluate the success of our use of Unity, we present evidence of versatility through the following selection of papers, published using environments that were created using our approach. 

\begin{itemize}
    \item \textit{Structured agents for physical construction} \citep{bapst2019structured}: This paper uses a continuous, procedurally-generated 2D world that tests an agent's ability to solve a variety of physics-based construction tasks.
    For each step, an agent can either choose from a selection of blocks to place in the scene, or glue existing blocks together in order to reach a variety of goals.
    Once an action is taken, the environment runs the simulation forward until all blocks come to rest.
    The episode terminates if the task is achieved, any of the blocks placed by the agent intersect with an obstacle or the maximum number of actions is exceeded.
    \item \textit{Making Efficient Use of Demonstrations to Solve Hard Exploration Problems} \citep{paine2019making}: Introduces a suite of eight hard exploration, first-person, 3D tasks with highly variable initial conditions.
    In each task, an agent must interact with objects in the world in order to gain access to a large apple that provides reward.
    \item \textit{Shaping Belief States with Generative Environment Models for RL} \citep{gregor2019shaping}: This paper utilizes three different environments, two of which are built using our approach:
    \begin{itemize}
        \item \textit{Random City}: a procedurally-generated, 3D navigation environment.
        At the start of each episode, a uniform distribution of colored boxes (i.e. "buildings") are placed atop a 2D plane, which is then used to generate training data to analyze the model's belief state (no reinforcement-learning in this experiment).
        \item \textit{Voxel Environment}: a voxel-based, procedural environment that can be modified by agents via building mechanisms. The environment consists of different block types that are placed in a fixed 3-dimensional grid. Some of these blocks can be picked up and placed by an agent, with the goal for each task variant to consume all yellow blocks, each of which gives a positive reward.
    \end{itemize} 
    \item \textit{Generalization of Reinforcement Learners with Working and Episodic Memory} \citep{fortunato2019generalization}: A suite of thirteen diverse machine-learning tasks that require memory to solve, of which eight were developed with our approach.
    Of those eight, four test whether the agent can correctly identify the difference between two nearly identical scenes, three are inspired by the Morris watermaze experiments \citep{miyake1999models}, and the last task tests if an agent can learn an overall transitive ordering over a chain of objects, through being presented with ordered pairs of adjacent objects.
    All of these tasks are set in a first-person, 3D world with some degree of variance in initial conditions.
\end{itemize}
\begin{figure}[H]
    \centering
	\includegraphics[width=0.40\columnwidth]{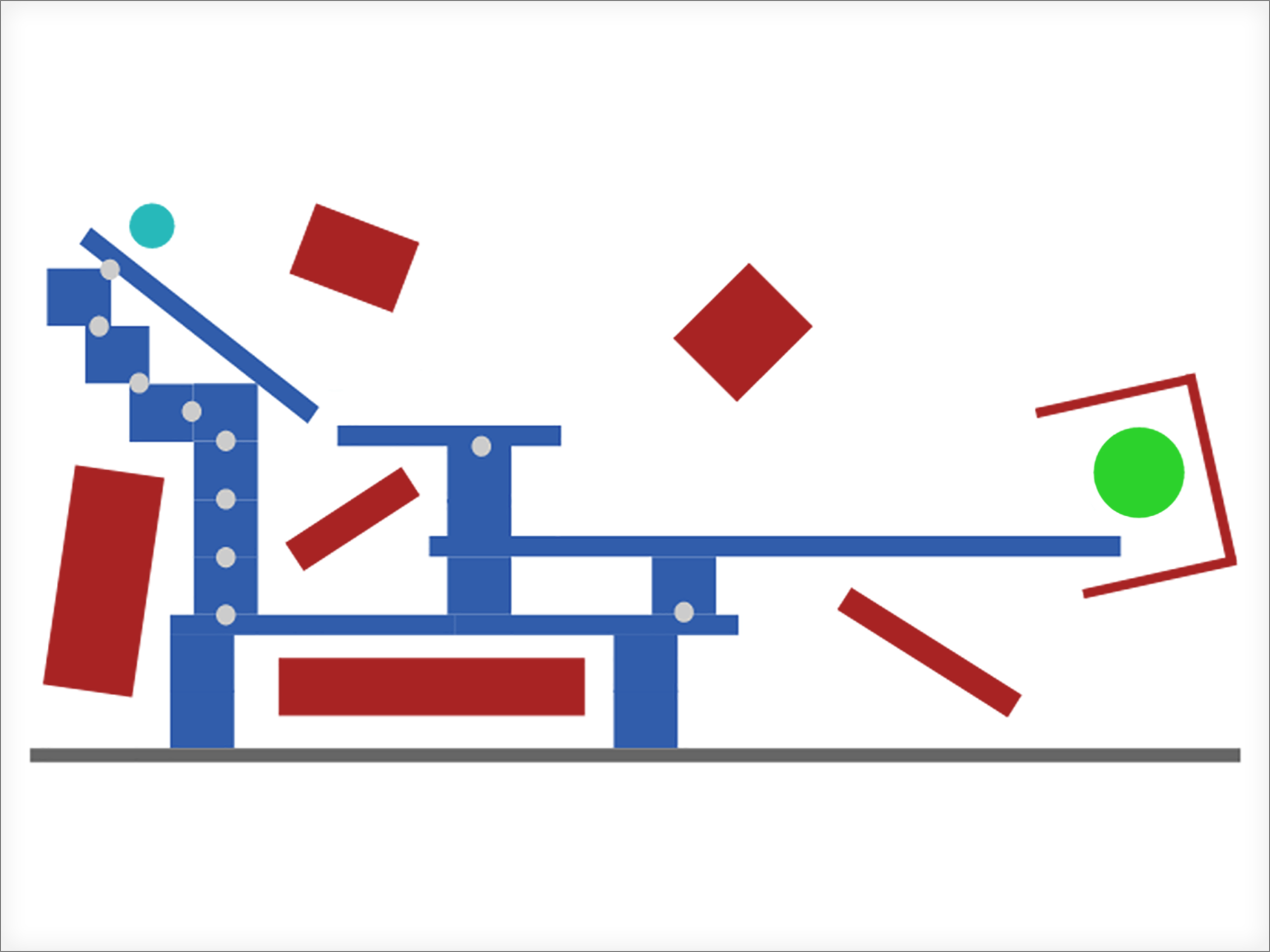}
	\includegraphics[width=0.40\columnwidth]{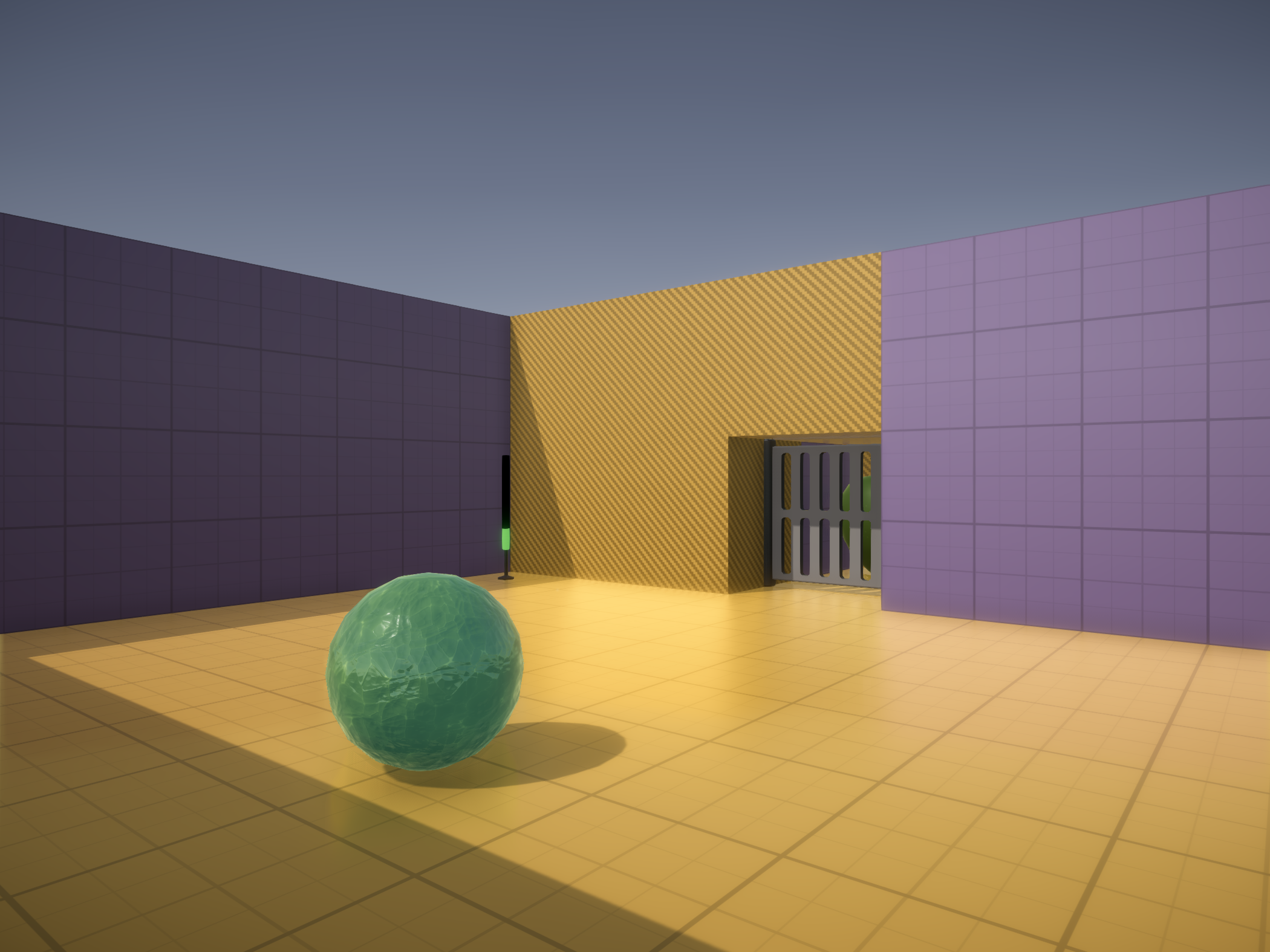}\\
	\includegraphics[width=0.40\columnwidth]{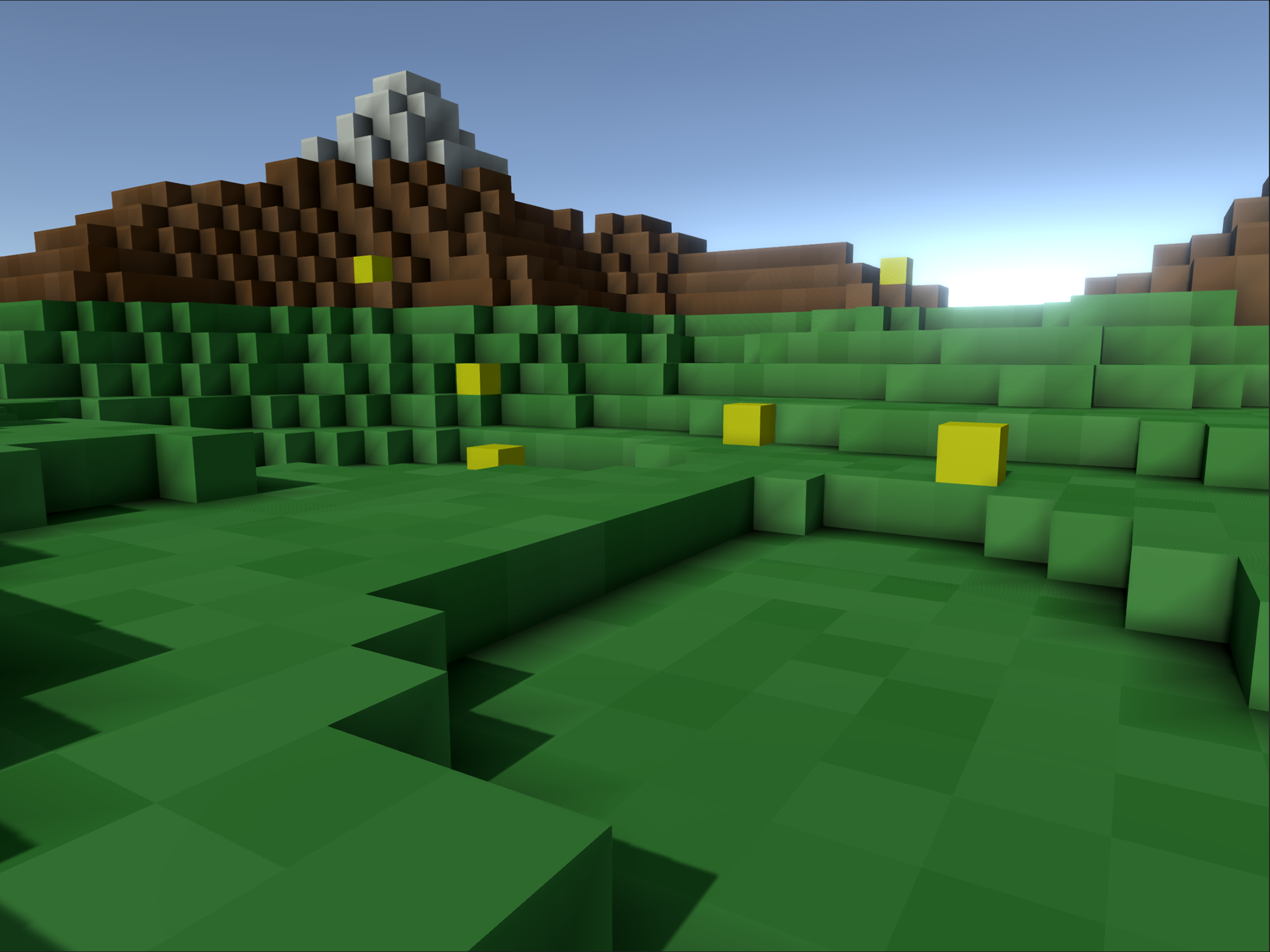}
	\includegraphics[width=0.40\columnwidth]{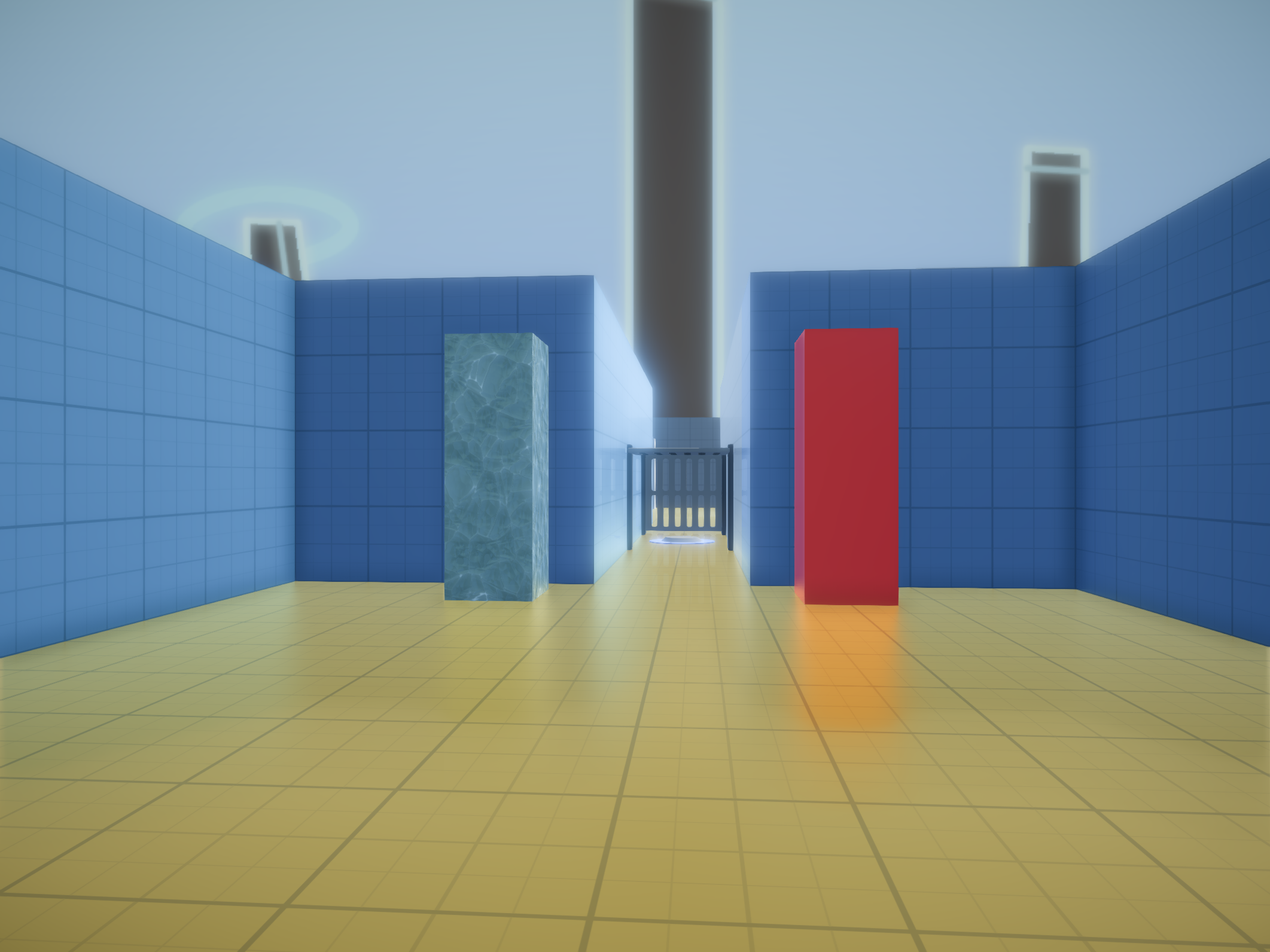}
	\caption{Screen captures from environments created in a selection of published papers. Clockwise from top left: 2-D construction task, hard exploration task, spot-the-difference memory task, procedurally-generated voxel-based task.}
    \label{fig:envs}
\end{figure}
\section{Conclusion}

We have presented how we use Unity to enable the creation of a large number of heterogeneous environments, for use in evaluating and developing artificial general intelligence.
Our use of Unity lets us leverage its flexibility in visual and simulation fidelity, and its programmer-friendly tools, to make it easier for researchers to author the environments necessary for evaluating their algorithmic advances.
We have described a means of packaging environments for distribution that enables the robust reproduction of results, and shown that the resultant environments are sufficiently computationally efficient for meaningful research to be performed.
Finally, we have evaluated the success of our use of Unity by presenting a selection of published papers that employ environments created using our approach.

We hope the concepts described in adapting Unity to our needs will aid others in the development of future simulated environments, using the open-sourced communication protocol and packaging methods to improve reproducible results.
\section{Acknowledgements}

This work would not have been possible without the support of our many colleagues at DeepMind.
In particular we would like to thank Robin Alazard, Pauline Coquinot, Tom Hudson, Jason Sanmiya and Marcus Wainwright for their contributions to this paper, Don Williamson for his technical expertise, and finally Tim Harley, Max Jaderberg and Patrick Pilarski for their insightful feedback.

We would also like to thank everyone at Unity Technologies for their continued support.
In particular, we would like to thank Vilmantas Balasevicius, Vincent-Pierre Berges, Arthur Juliani, Danny Lange, Marwan Mattar, and Jeffrey Shih.

\bibliographystyle{abbrvnat}
\setlength{\bibsep}{5pt} 
\nobibliography*
\bibliography{main.bbl}

\end{document}